\documentclass[letterpaper, 10pt, conference]{ieeeconf}
\IEEEoverridecommandlockouts  
\usepackage{color}
\usepackage{enumerate}
\usepackage{algorithm}
\usepackage{algorithmic}
\usepackage{multicol}
\usepackage{amssymb, amsmath}
\usepackage{stmaryrd}
\usepackage{graphicx}
\usepackage{setspace}
\usepackage{mathrsfs}
\usepackage{bm}
\usepackage[bookmarks=true]{hyperref}

\newtheorem{prop}{Property}

\def\bfzero{\mathbf{0}}

\def\DS{\textrm{DS}}
\def\HOUBA{\textrm{HOUBA}}
\def\OGH{\textrm{OGH}}
\def\SS{\textrm{SS}}
\def\cur{\textrm{cur}}
\def\eg{\emph{e.g.}}
\def\goal{\textrm{goal}}
\def\ie{\emph{i.e.},~}

\def\pdd{\ddot{\bfp}}
\def\pd{\dot{\bfp}}
\def\pss{{\bfp}_{ss}}
\def\ps{{\bfp}_s}

\def\qd{\dot{\q}}
\def\q{\bfq}
\def\sdd{\ddot{s}}
\def\sddmax{\ddot{s}_{\textrm{max}}}
\def\sd{\dot{s}}

\def\swing{\textrm{swing}}
\def\trans{\textrm{trans}}

\newcommand{\bfa}{{\bm{a}}}
\newcommand{\bfb}{{\bm{b}}}
\newcommand{\bfc}{{\bm{c}}}

\newcommand{\bff}{{\bm{f}}}
\newcommand{\bfg}{{\bm{g}}}

\newcommand{\bfn}{{\bm{n}}}

\newcommand{\bfp}{{\bm{p}}}
\newcommand{\bfq}{{\bm{q}}}

\newcommand{\bft}{{\bm{t}}}

\newcommand{\bfv}{{\bm{v}}}
\newcommand{\bfw}{{\bm{w}}}
\newcommand{\bfx}{{\bm{x}}}

\newcommand{\bfA}{\mathbf{A}}
\newcommand{\bfB}{\mathbf{B}}

\newcommand{\bfL}{\mathbf{L}}

\newcommand{\calP}{{\cal P}}

\newcommand{\colvec}[1]{\begin{bmatrix} #1 \end{bmatrix}}
\newcommand{\defeq}{\stackrel{\mathrm{def}}{=}}
\renewcommand\d[1]{{\rm d}{#1}}
\renewcommand{\th}[1]{\ensuremath {#1}^{\textrm{th}}}

\title{When to make a step? Tackling the timing problem in multi-contact
locomotion by TOPP-MPC}

\author{St\'{e}phane Caron$^{1}$ and Quang-Cuong Pham$^{2}$%
    \thanks{$^{1}$ Laboratoire d'Informatique, de Robotique et de
    Micro\'{e}lectronique de Montpellier (LIRMM), CNRS--University of
    Montpellier, France.}%
    \thanks{$^{2}$ School of Mechanical and Aerospace Engineering,
    Nanyang Technological University, Singapore.}%
    \thanks{* This work is supported in part by H2020 EU project COMANOID
    \url{http://www.comanoid.eu/}, RIA No 645097. \newline
    Corresponding author: {\tt\footnotesize stephane.caron@normalesup.org}}%
}

\begin{document}

\maketitle

\begin{abstract}
    We present a model predictive controller (MPC) for multi-contact locomotion
    where predictive optimizations are realized by time-optimal path
    parameterization (TOPP). A key feature of this solution is that, contrary
    to existing planners where step timings are provided as inputs, here the
    timing between contact switches is computed as \emph{output} of a fast
    nonlinear optimization. This is appealing to multi-contact locomotion,
    where proper timings depend on terrain topology and suitable heuristics are
    unknown. We show how to formulate legged locomotion as a TOPP problem and
    demonstrate the behavior of the resulting TOPP-MPC controller in
    simulations with a model of the HRP-4 humanoid robot.
\end{abstract}

\section{Introduction}

Walking pattern generators on horizontal floors usually take single-support
(SS) and double-support (DS) durations as user-defined parameters. Recent works
showed how adapting step timings helps recover from
disturbances~\cite{khadiv2016step, griffin2017walking}, yet using capture-point
schemes that are so far limited to walking on flat surfaces. The full
locomotory capabilities of humanoids appear in \emph{multi-contact}, where
wheeled robots cannot follow. In multi-contact locomotion, walking speed and
step timings depend on terrain topology, as illustrated for instance by
Naismith's rule: the walking \emph{pace} (in min$.$km$^{-1}$) depends affinely
on terrain slope. 

Model predictive control (MPC) is a paradigm that gives controllers the level
of foresight required to tackle the issue of timed walking. Its application to
multi-contact locomotion is relatively recent, and can be split in two
categories. In one line of research, contact forces are jointly considered as
control variables used to optimize a cost function over future whole-body
motions~\cite{audren2014iros, herzog2015humanoids, carpentier2016icra}. In such
formulations, contact feasibility constraints (namely, that contact wrenches
lie inside their wrench cone) are straightforward to calculate, at the cost of
a large number of control variables.

Another line of research seeks to reduce both control variables and contact
constraints at the center of mass (COM), \ie focusing on \emph{centroidal
dynamics}~\cite{wensing2016ijhr}. In~\cite{caron2017zmp}, ZMP support areas
were generalized to multi-contact and applied to whole-body motion generation,
but it was observed that these areas vary with the position of the COM. This is
indeed a general phenomenon: once reduced at the COM, contact feasibility
constraints yield \emph{bilinear} inequalities crossing COM position and
acceleration. In~\cite{brasseur2015humanoids}, these inequalities were kept
linear by bounding variations of a nonlinear term, resulting in a ZMP
controller that can raise or lower its COM. Polyhedral boundaries were also
used in~\cite{caron2016multi} to formulate a linear MPC problem over 3D
accelerations of the COM.

Yet, all of the works~\cite{audren2014iros, herzog2015humanoids, caron2017zmp,
brasseur2015humanoids, caron2016multi, vanheerden2017ral, naveau2017ral} are
based on pre-defined timings. Set aside flat-floor walking, the alternative to
fixed timings seems to be blackbox nonlinear
optimization~\cite{mordatch2010tog, lengagne2013ijrr, carpentier2016icra}.
Walking on non-flat terrains was showed in~\cite{mordatch2010tog} using a SLIP
model and on-line foot-step planning. Lengagne \emph{et
al.}~\cite{lengagne2013ijrr} showcased a broad set of tasks, but noted that the
performance and numerical stability of nonlinear solvers were still
problematic. Recent developments~\cite{naveau2017ral, vanheerden2017ral}
performed a few iterations of sequential quadratic programming fast enough for
the control loop; yet again using pre-defined step timings.

In this work, we explore an alternative to linearize bilinear constraints at
the center of mass: namely, by alternating path interpolation with trajectory
retiming. A key feature of this approach is that all timings (step durations,
COM transfer durations, etc.) are produced as \emph{output} of the optimization
problem. Our contribution lies in (1) the formulation of legged locomotion as a
TOPP problem, including COM-trajectory retiming, contact switches and swing-leg
synchronization, and (2) a model predictive controller (called TOPP-MPC) that
leverages this solution into a full-fledge multi-contact locomotion controller.

\section{Background}
\label{sec:background}

\subsection{Contact stability conditions}

We wish that contacts stay \emph{fixed} (\ie neither slip nor tilt) while the
robot pushes on them to locomote. The inequality constraints that model this
regime are known as \emph{contact-stability} conditions, and can be rewritten
at the centroidal level. Consider the Newton-Euler equations of motion:
\begin{equation}
  \label{eq:EOM}
  \colvec{m\ddot\bfp_G\\ \dot \bfL_G} = \colvec{m\bfg \\ \bfzero} +
  \bfw_G, 
\end{equation}
where $m$ is the total robot mass, and $G$ its center of mass (COM) located at
$\bfp_G$ (coordinates are given in the inertial frame of origin $O$), $\bfL_G$
is the angular momentum of the robot around $G$, $\bfg$ the gravity vector and
$\bfw_G$ the net contact wrench taken at $G$:
\[
\bfw_G := \sum_{i=0}^K\colvec{\bff_i\\ \overrightarrow{GC_i} \times \bff_i},
\]
where $\bff_i$ is the contact force exerted onto the robot at the
$i^\mathrm{th}$ contact point $C_i$ (this formulation includes surface
contacts with continuous pressure distributions, see \eg~\cite{caron2015icra}).
We can rewrite the Newton-Euler equations~\eqref{eq:EOM} at a fixed reference
point $O$ as:
\begin{equation}
  \label{eq:EOM2}
  \bfw_O \ = \ \colvec{m(\ddot\bfp_G-\bfg)\\ \dot \bfL_G + m\bfp_G \times
    (\ddot\bfp_G-\bfg)}.
\end{equation}
Under the assumption of linearized friction cones, \emph{feasible} contact
wrenches, \ie those corresponding to contact forces that lie inside their
respective friction cones, are exactly characterized by:
\begin{equation}
  \label{eq:CWC}
  \bfA_O \bfw_O \leq \bfzero,
\end{equation}
where $\bfA_O$ is the matrix of the net contact wrench cone (CWC), which can be
computed based uniquely on the positions and orientations of the contacts. See
\emph{e.g.}~\cite{caron2015rss} for a review of the corresponding algorithm.

\subsection{TOPP and TOPP-Polygon}

Consider a robot with $n$ degrees of freedom and a path $\bfq(s)_{s \in [0,
1]}$ in its configuration space. Assume that all dynamic constraints on the
robot along the path can be expressed in the form:
\begin{equation}
  \label{eq:gen}
  \ddot s \bfa(s) + \dot s^2 \bfb(s) + \bfc(s) \leq 0.
\end{equation}
Finding the time-optimal parameterization $s(t)_{t \in [0, T]}$ subject to
constraints~(\ref{eq:gen}) is the classical time-optimal path parameterization
(TOPP) problem, for which efficient methods have been developed
(see~\cite{Pha14tro} for a historical review).

A wide range of constraints can be put into the form of~(\ref{eq:gen}),
including pure acceleration bounds, torque bounds for serial manipulators,
contact-stability constraints for humanoid robots in
single-~\cite{caron2015icra} and multi-contact~\cite{caron2015rss}. Constraints
on redundantly actuated systems (such as humanoids in multi-contact), cannot be
put into the form of~\eqref{eq:gen}, but rather in the more general
form~\cite{Hau14ijrr}:
\begin{equation}
    \label{eq:over}
    (\ddot s , \dot s^2) \in \calP(s),
\end{equation}
where $\calP(s)$ is a convex polygon in the $(\ddot s,\dot s^2)$-plane.
In~\cite{PS15tmech}, the authors developed TOPP-Polygon, an extension of the
TOPP algorithm that is able to deal with such polygonal constraints. Its
computation bottleneck lied in the enumeration of polygonal constraints
$\calP(s)$, which could take up to tens of milliseconds per path discretization
step. In what follows, we reduce it to less than a millisecond.

\section{TOPP for multi-contact locomotion}
\label{sec:TOPP}

\subsection{Reduction of TOPP Polygons}
\label{reduction}

Consider a path $\bfp_G(s)_{s\in[0, 1]}$ of the center of mass. Differentiating
twice, one obtains
\begin{equation}
    \ddot\bfp_G = {\bfp_G}_s \ddot s + {\bfp_G}_{ss}\dot s^2,
\end{equation}
where the subscript $s$ denotes differentiation with respect to the
path parameter.

Assume that the angular momentum at the center of mass is kept constant
($\dot{\bfL}_G=\bfzero$). Substituting the expression of $\ddot{\bfp}_G$ into
the equation of motion~(\ref{eq:EOM2}), we get:
\begin{equation}
    \bfw_O = \colvec{
    m ({\bfp_G}_s \ddot s + {\bfp_G}_{ss}\dot s^2 - \bfg) \\
    {\bfp_G} \times m ({\bfp_G}_s \ddot s + {\bfp_G}_{ss} \dot s^2 -
    \bfg)}.
\end{equation}
The contact-stability condition~(\ref{eq:CWC}) thus rewrites to:
\begin{equation}
    \small
    \label{eq:A}
    \ddot s \bfA_O\colvec{{\bfp_G}_s\\ {\bfp_G} \times {\bfp_G}_s}
    + \dot s^2 \bfA_O\colvec{
        {\bfp_G}_{ss}\\ {\bfp_G}\times{\bfp_G}_{ss}}
    \leq \bfA_O\colvec{ \bfg\\ {\bfp_G}\times\bfg}
\end{equation}
which has the canonical form~(\ref{eq:gen}) of TOPP. 

Equation~(\ref{eq:A}) usually contains a large number of inequalities, up to
150 for a rectangular double-support area. Existing TOPP
solvers~\cite{Pha14tro, Hau14ijrr}) take several seconds to solve problems with
that many inequalities, which would make them unpractical for closed-loop
control. Hence the importance of pruning redundant
inequalities~\cite{Hau14ijrr}, which is algorithmically equivalent to applying
a convex hull to the dual of the problem, as exploited
in~\cite{caron2016multi}. 

This observation applies to the present TOPP problem. Equation~(\ref{eq:A}) can
be put in the canonical form:
\begin{equation}
    \bfB [\sdd\  \sd^2]^\top \leq \bfc
\end{equation}
where the matrix $\bfB$ and vector $\bfc$ are defined by:
\[
    \footnotesize
    \bfB, \bfc :=  \bfA_O \begin{bmatrix}
        {\bfp_G}_s & {\bfp_G}_{ss} \\
        {\bfp_G} \times {\bfp_G}_s & {\bfp_G}\times{\bfp_G}_{ss}
    \end{bmatrix},
\bfA_O\colvec{ \bfg\\ {\bfp_G}\times\bfg}.
\]
If the polygon thus described contains the origin $(0, 0)$ (\ie if $\bfc \geq
\bm0$) then running a convex hull algorithm on the rows of $\bfB$ (dual
vectors) will enumerate edges of the corresponding polygon. Intersecting
consecutive edges will then provide the list of vertices. Meanwhile, if $(0,
0)$ is outside of the polygon, it becomes necessary to compute an interior
point of the polygon. Following~\cite{caron2016multi}, we use its
\emph{Chebyshev center} $\mathring{\bfx}$, which can be computed by solving a
single linear program.

\begin{table}[t]
    \caption{
        Polygon sizes and computation times for polygon reduction by convex
        hull (CH) and Bretl \& Lall's method (B\&L). 
    }
    \centering
    \small
    \begin{tabular}{ccccc}
        Contact  & \# ineq. bef. & \# ineq. aft. & Hull$^1$ & B\&L$^1$ \\
        \hline   
        Single (SE) & 16 & 3.8 $\pm$ 0.4  & 0.4\,ms & 1.0\,ms \\
        Single (NSE) & 16 & 3.5 $\pm$ 0.5 & 0.6\,ms & 1.2\,ms \\
        Double & 145 & 6.1 $\pm$ 1.6 & 0.5\,ms & 2.0\,ms \\       
        \hline
    \end{tabular}
    \label{tab:polygons}
\end{table}

Alternatively, one can compute the polygon directly from~(\ref{eq:A}) using a
recursive polytope projection method~\cite{bretl2008tro}.
Table~\ref{tab:polygons} reports an experimental comparison of the two
methods\footnote{All computation times reported in this paper were obtained on
an Intel\textsc{(r)} Core\textsc{(tm)} i7-6500U CPU @ 2.50 Ghz. Standard
deviations for computation times in Table~\ref{tab:polygons} are not reported
as they are all smaller than 0.1\,ms.} (convex hull versus polytope
projection), in three different scenarios: single-contact in static equilibrium
(SE), single-contact non-static-equilibrium (NSE) and double contact. The table
reports the number of inequalities before and after pruning, averaged across
multiple positions $\bfp_G$, with computation times averaged across 100
iterations. We observe that the convex-hull method outperforms the polytope
projection method in all scenarios, and significantly so in double contact.

\begin{figure}[t]
    \centering    
    \includegraphics[width=0.97\columnwidth]{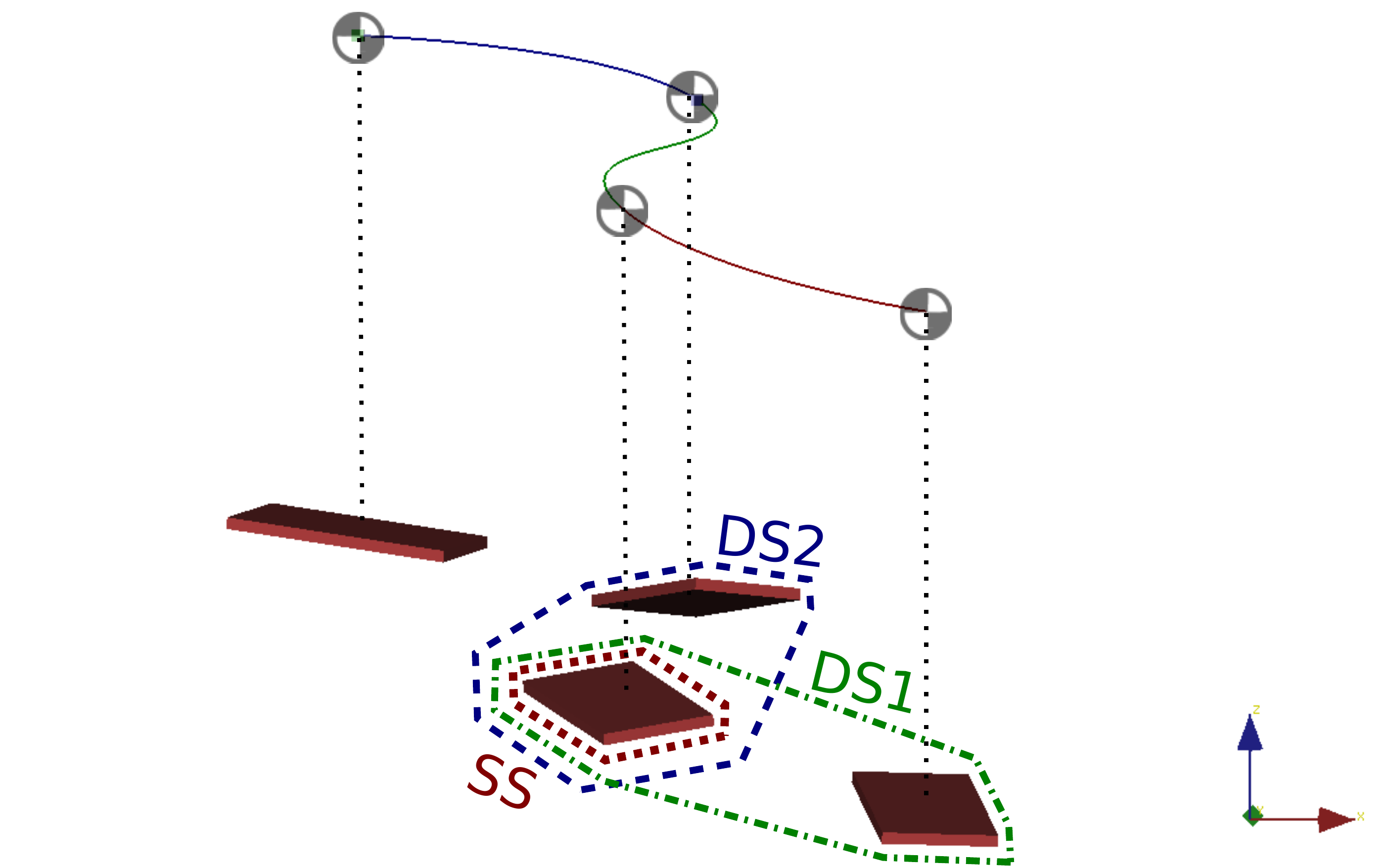}
    \caption{
        Contact locations and interpolated COM paths for the test case reported
        in Table~\ref{tab:polygons}. Contacts corresponding to the successive
        Double-Support (DS) and Single-Support (SS) phases are contoured in
        dotted lines.
    }
    \label{fig:3D}
\end{figure}

\subsection{TOPP through contact switches}
\label{sec:multi}

\begin{figure}[t]
    \centering
    $s=0.4$ (DS1) \hspace{2cm} $s=0.7$ (SS)\\
    \includegraphics[width=0.23\textwidth]{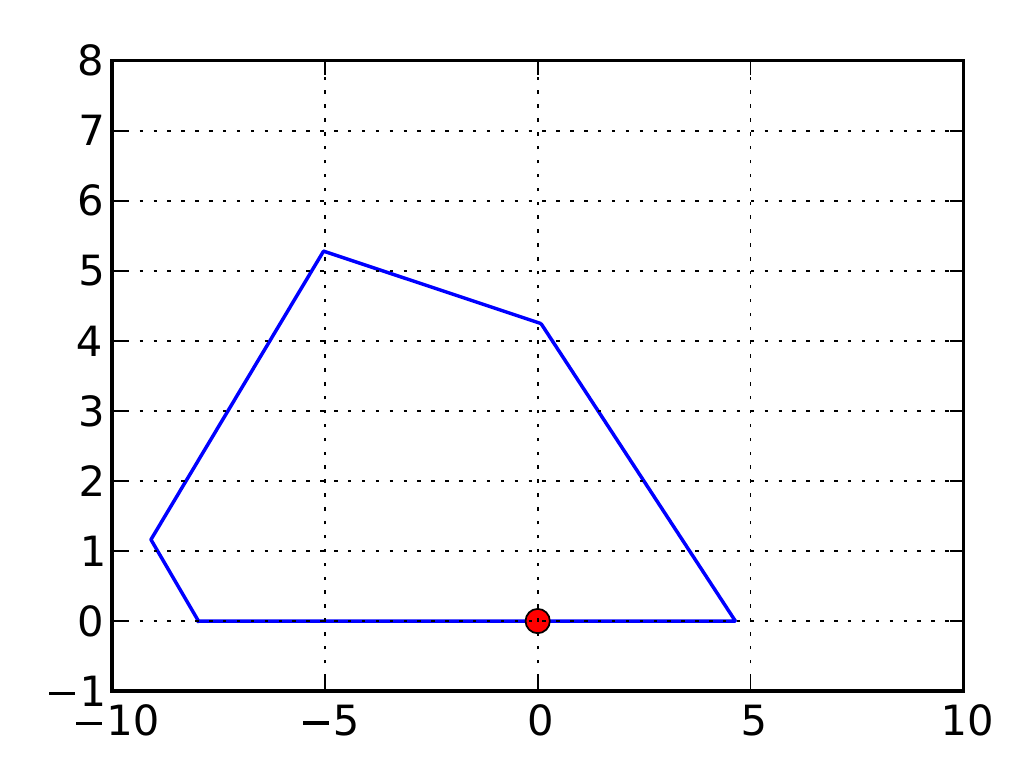}
    \includegraphics[width=0.23\textwidth]{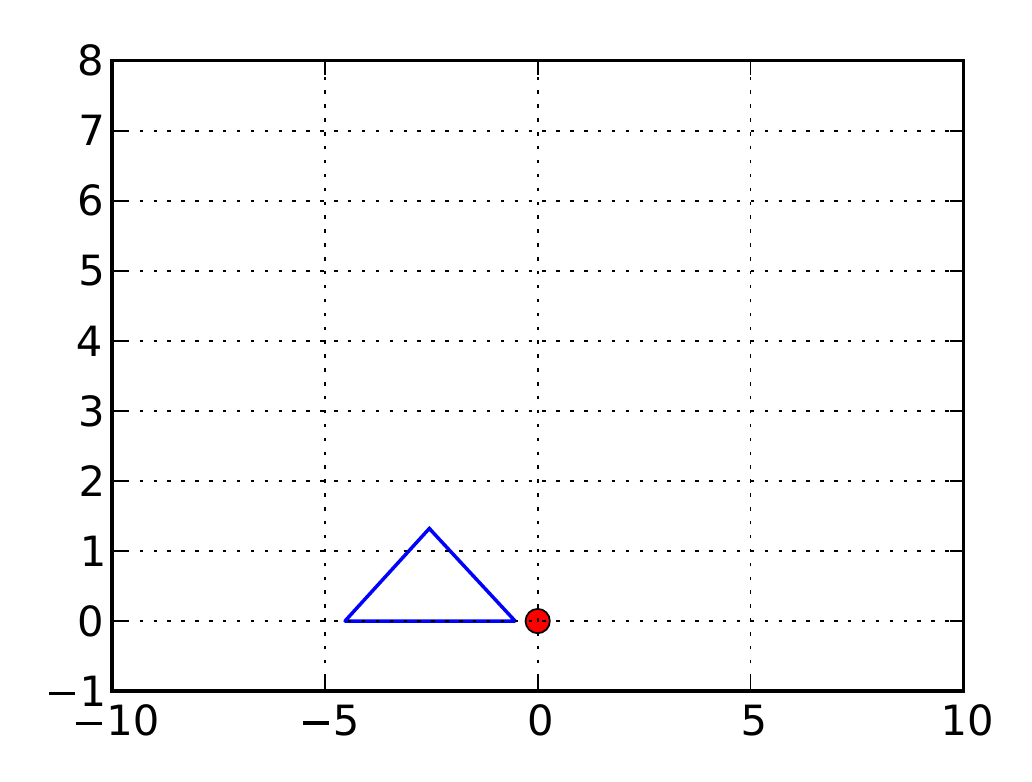}\\
    $s=0.76$ (SS) \hspace{2cm} $s=1.45$ (DS2)\\
    \includegraphics[width=0.23\textwidth]{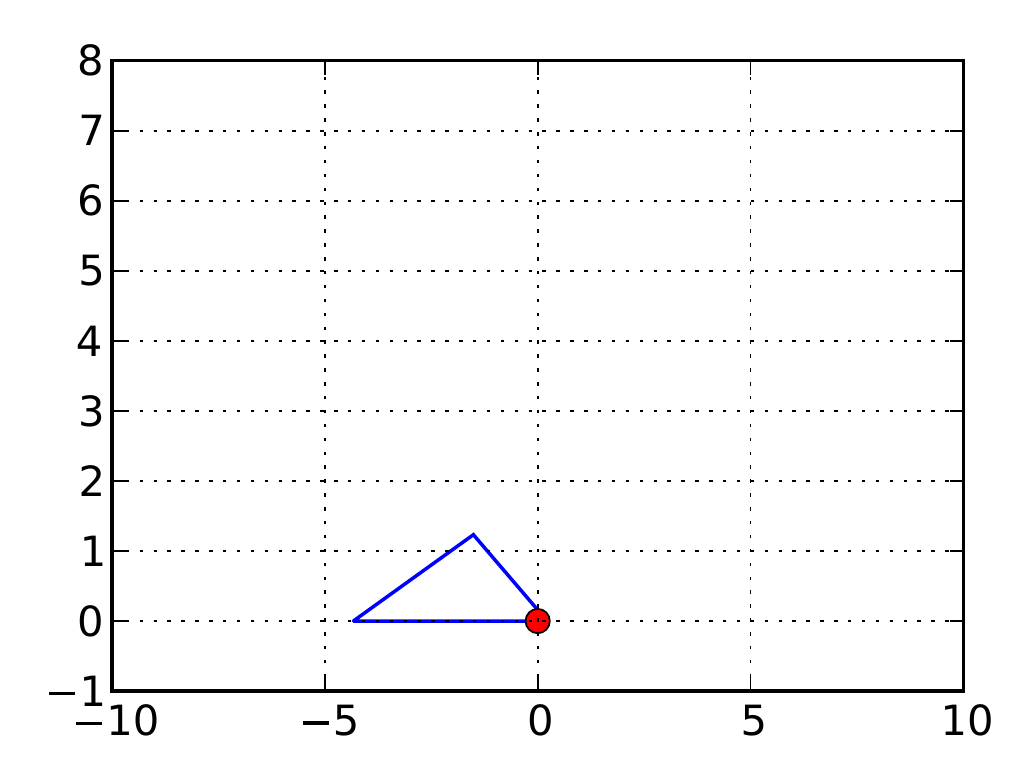}
    \includegraphics[width=0.23\textwidth]{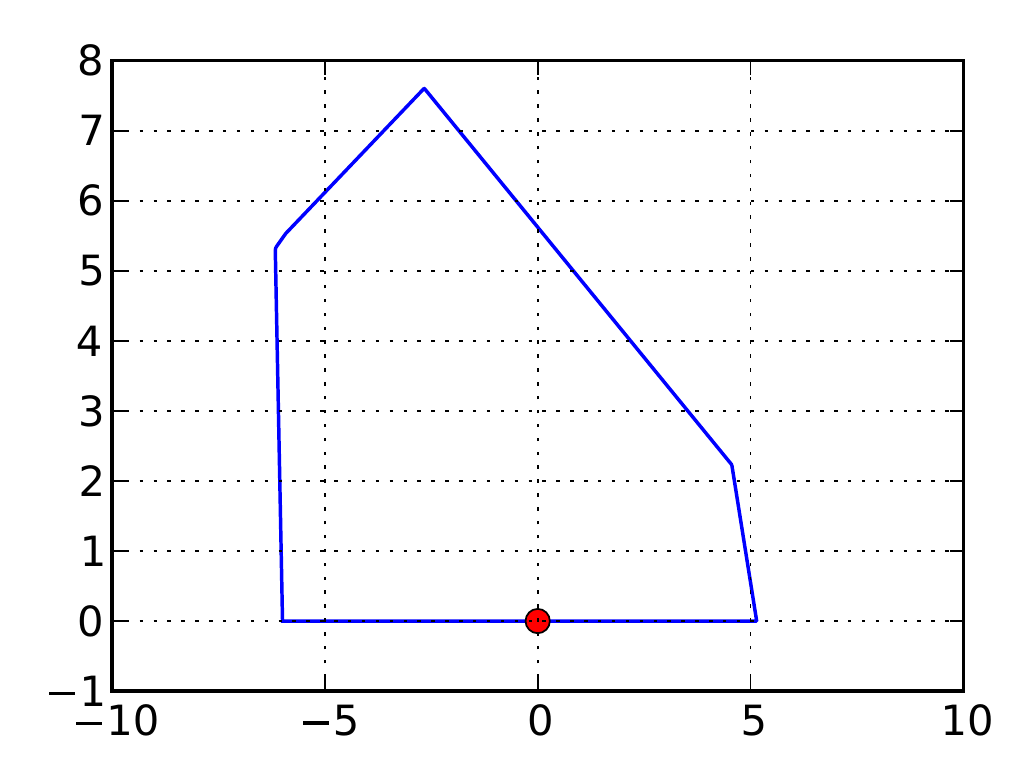}\\
    \caption{
        Constraint polygons in the $(\ddot s,\dot s^2)$-plane for various
        values of $s$. Here, the contact switches happen at $s_1=0.67$ (DS1 to
        SS) and $s_2=1.33$ (SS to DS2). Note that the beginning of the
        single-contact phase is not in static equilibrium, as illustrated by
        the polygon at $s=0.7$ which does not contain the origin (red dot).
    }
    \label{fig:polygons}
\end{figure}

Consider the smooth three-dimensional COM path ${\bfp_G}(s)_{s\in[0,1]}$
depicted in Figure~\ref{fig:3D}, along with the sequence of foot stances
DS1--SS--DS2. Denote by $s_1$ (resp. $s_2$) the path index of the contact
switch from DS1 to SS (resp. from SS to DS2). Mind that $s_1$ and $s_2$ are two
geometric positions and do not include any timing information. Consider now the
TOPP problem \emph{through contact switches}. The contact-stability matrices
$\bfA_{\DS1}$, $\bfA_{\SS}$, $\bfA_{\DS2}$ (computed only once per stance based
on footstep locations~\cite{caron2015rss}) are used over the path intervals
$[0,s_1]$, $[s_1,s_2]$, $[s_2, 1]$ respectively. Figure~\ref{fig:polygons}
shows examples of polygons computed in each of the intervals. TOPP can finally
be run on the full path $[0, 1]$, subject to the constraints provided by these
polygons. 



\subsection{Geometric nature of contact switches}
\label{sec:nqs}


Single-support phases allow bipeds to transfer one foot (the swing foot) to a
new foothold location while the other (the support foot) stays fixed. A central
question is then: how should the COM move during these phases? One answer is to
maintain it inside the static-equilibrium prism (SEP) of the support
foot,\footnote{
    When walking on horizontal floors, this is equivalent to keeping the COM
    over the so-called support area.
} a behavior that can be observed in many walking patterns reported in the
literature (see \eg~Figure~4 in~\cite{buschmann2012humanoids}, Figure~1
in~\cite{santacruz2012iros}, Figure~2 and 3 in~\cite{vanheerden2017ral}, ...).
However, having the COM in the SEP indicates \emph{quasi-static} walking. It
notably precludes the discovery of dynamic walking patterns, where the COM need
not enter the SEP if foot swings are fast enough. TOPP can discover such
motions through proper variations of the geometric parameters $s_1$ and $s_2$.

Let $s^*_1$ and $s^*_2$ denote respectively the first and second intersection
of the COM path with the static-equilibrium prism.
Choosing $s_1=s^*_1$ and $s_2=s^*_2$ corresponds to the ``safe'' quasi-static
option. By contrast, choosing $s_1 < s^*_1$ and $s_2 > s^*_2$ gives
rise to portions of the path that are not quasi-statically traversable.
However, if the path is time-parameterizable, there will exist a
dynamically-stable execution. Furthermore, as $s_1$ gets smaller and and $s_2$
gets larger, the time-parameterized behavior will be more aggressive, up to a
point when no a feasible time-parameterization exists. For the same path as in
Figure~\ref{fig:3D}, if one chooses $s_1=0.51$ instead of $s_1=0.67$, then the
path becomes non time-parameterizable.

\section{Locomotion by Retimed Predictive Control}
\label{sec:MPC}

The loop of a predictive controller can be described as follows: at each
iteration, (1) generate a \emph{preview trajectory} of future system dynamics
leading to a desired configuration, then (2) apply the \emph{first controls} of
this trajectory. To keep up with the high rates of a control loop, the preview
trajectory is commonly found as the solution to an optimization problem, be it
a linear-quadratic regulator~\cite{kajita2003icra, tedrake2015humanoids}, a
quadratic program~\cite{audren2014iros, caron2016multi} or a blackbox
optimal-control problem~\cite{carpentier2016icra}. In what follows, our
optimization unfolds in two steps: interpolating preview paths, and retiming
them using TOPP. Our preview window foresees one footstep ahead at a time.

In single-support phases, one needs to interpolate both COM and swing-foot
trajectories. Similarly to \cite{takenaka2009iros}, we adopt a simplified
dynamics model where the COM is represented by a point-mass and the swing foot
by a rigid body (Figure~\ref{fig:synchro}). Contact dynamics are reduced at the
COM under zero angular momentum as described in Section~\ref{reduction}, while
joint kinematic and dynamic constraints are modeled by workspace velocity and
acceleration limits on the swing foot. All of these constraints are taken into
account when retiming by TOPP. Retimed foot and COM accelerations are then sent
as reference to a whole-body controller that converts them into joint
accelerations. The complete pipeline is summarized in
Figure~\ref{fig:pipeline}.

\subsection{Finite state machine}

The locomotion state machine underlying TOPP-MPC is simpler than those of
previous works like~\cite{caron2016multi} as it is does not need to take time
into consideration. Transitions between single- and double-support phases are
triggered by straightforward geometric conditions, namely:
\begin{itemize}
    \item SS $\to$ DS when $\|\bfp_\swing - \bfp_\swing^\goal\| \leq \epsilon$,
        \ie when the swing-foot touches down on its target foothold.
    \item DS $\to$ SS when $\|\bfp_G - \bfp_G^\goal\| \leq d_\trans$, \ie when
        the COM enters the vicinity of its preview target. 
\end{itemize}
While $\epsilon$ should be a small value scaled upon the ability of the robot
to estimate its foot displacements, $d_\trans$ depends on the contact geometry
and preview target velocity. We used $d_\trans = 5$ cm in our experiments,
which corresponds roughly to the half-width of static-equilibrium polygons.

\subsection{COM and swing-foot path interpolation}

Path interpolation is constrained by state estimation from the robot: the
beginning of the path $\bfp(s)$ needs to coincide with the current robot state
$(\bfp_\cur, \pd_\cur)$, where we use $\bfp$ to refer equivalently to the COM
position $\bfp_G$ or swing-foot position $\bfp_\swing$. This means that:
\begin{eqnarray}
    \bfp(0) & = & \bfp_\cur \label{path-p0} \\
    \cos{(\ps(0), \pd_\cur)} 
    &= & 1 \label{path-v0}
\end{eqnarray}
The latter equation, a cosine of the angle between two vectors, states that the
initial path tangent must be positively aligned with the current velocity. The
norms of the two vectors will be matched as well at the retiming stage, so that
the output trajectory velocity $\pd(0) = \ps(0) \sd(0) = \pd_\cur$.

\begin{figure}
    \includegraphics[width=0.9\columnwidth]{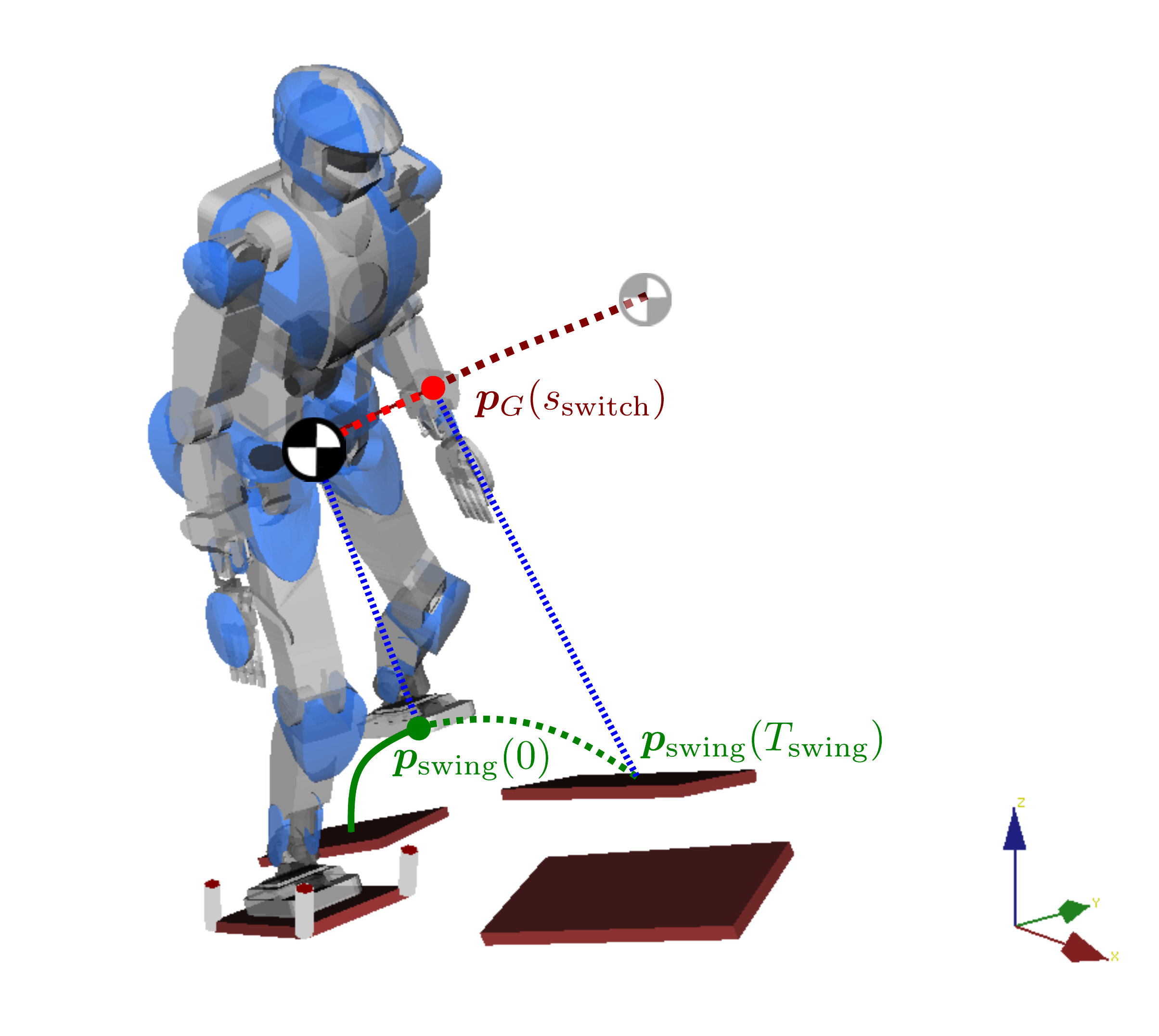}
    \caption{
        \textbf{Preview of swing-foot (green) and COM (ligh and dark red)
        trajectories during a single-support phase.} Retiming the swing foot
        path under conservative constraints yields a swing duration $T_\swing$.
        This value is then converted into retiming constraints on the COM
        trajectory, so that TOPP uses the single-support CWC up to
        $\bfp_G(s_1)$ (light red) and the double-support one afterwards (dark
        red).
    }
    \label{fig:synchro}
\end{figure}

Target positions and velocities $(\bfp_\goal, \pd_\goal)$ by the end of the
preview window are also provided:
\begin{itemize}
    \item for swing-foot paths, the target position is taken at the contact
        location and the target velocity is zero;
    \item for COM paths, the target position is taken at the center of the
        Static-Equilibrium Prism for the next single-support phase, while the
        target velocity is a fixed-norm vector parallel to the contact surface
        and going forward in the direction of motion.
\end{itemize}
Boundary conditions at the end of the preview path are similarly given by:
\begin{eqnarray}
    \bfp(1) & = & \bfp_\goal \label{path-p1} \\
    \cos(\ps(1), \pd_\goal)
    &= & 1 \label{path-v1}
\end{eqnarray}

\begin{figure*}
    \includegraphics[width=\textwidth]{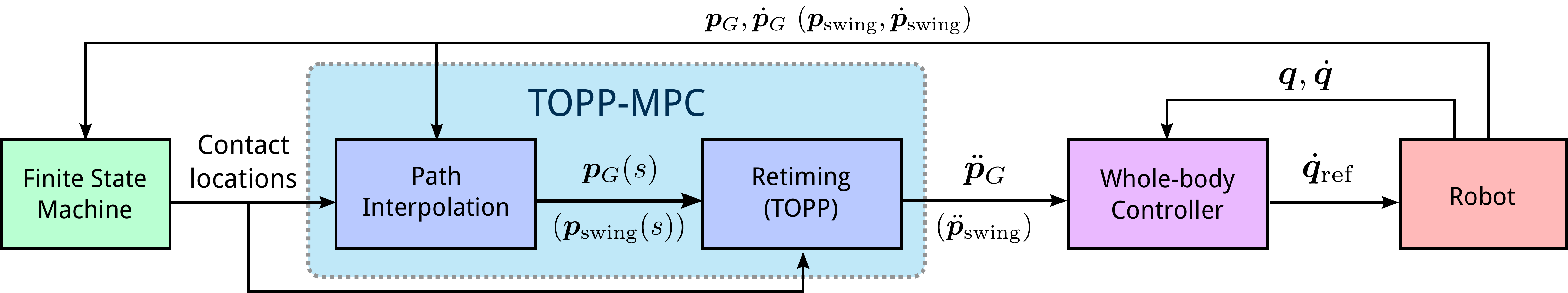}
    \caption{
        \textbf{Pipeline of the predictive controller.} A finite state machine
        sends the current walking phase (single- or double-support) as well as
        contact locations to TOPP-MPC. COM and (in single support) swing-foot
        trajectories are then interpolated from the current robot state to a
        desired future state, and sent to TOPP. The initial COM and foot
        accelerations of the retimed trajectory are finally converted to joint
        accelerations by a whole-body controller and sent to the robot.
    }
    \label{fig:pipeline}
\end{figure*}

Like~\cite{Hau14ijrr}, we interpolate the path $\bfp(s)$ by a cubic Hermite
curve, \ie a third-order polynomial $H(\bfp_0, \bfv_0, \bfp_1, \bfv_1)$ such
that $H(0)=\bfp_0$, $H'(0)=\bfv_0$, $H(1)=\bfp_1$ and $H'(1)=\bfv_1$. This
construction ensures that position constraints \eqref{path-p0} and
\eqref{path-p1} are satisfied. However, velocity constraints \eqref{path-v0}
and \eqref{path-v1} are not vector equalities: they only impose vector directions
and signs, leaving norms as a free parameter. Our path interpolation problem
therefore has two degrees of freedom $\lambda > 0$ and $\mu > 0$: we can choose
any path
\begin{equation}
    I_{\lambda, \mu} = 
    H(\bfp_\cur,\,\lambda \pd_\cur,\,\bfp_\goal,\,\mu \pd_\goal)
\end{equation}

The values of $\lambda$ and $\mu$ can be selected so as to optimize additional
path smoothness criteria. This problem has been studied in computer graphics,
where Yong et al.~\cite{yong2004cagd} introduced the Optimized Geometric
Hermite (OGH) curves that minimize \emph{strain energy} $\int_0^1 \|\pss(s)\|^2
\d{s}$ of the path:
\begin{equation}
    \label{eq:OGH}
    \OGH(\bfp_0, \bfv_0, \bfp_1, \bfv_1) \ = \ \underset{\lambda,
    \mu}{\arg\min} \int_0^1 \|I_{\lambda, \mu}''(s)\|^2 \d{s}
\end{equation}
An interesting feature of this problem is
that the values $\lambda^*_\OGH, \mu^*_\OGH$ that yield this minimum are found
analytically from boundary conditions $(\bfp_\cur, \pd_\cur, \bfp_\goal,
\pd_\goal)$, so that there is no need for numerical optimization at runtime. We
experimented with OGH curves but observed that, when boundary velocity vectors
$\pd_\cur$, $\pd_\goal$ deviate significantly from the direction of motion
$\Delta \defeq \bfp_\goal - \bfp_\cur$, OGH curves tend to start or end with very
sharp accelerations. These accelerations have little impact on strain energy
but jeopardize trajectory retiming.

To avoid this phenomenon, we chose to optimize a different criterion. We
observed empirically that a good proxy criterion for the ``retimability'' of a
trajectory is the uniform bound given by:
\begin{equation}
    \label{eq:UOGH}
    \underset{\lambda, \mu}{\arg\min} \max_{s \in [0, 1]} \|I_{\lambda, \mu}''(s)\|^2
\end{equation}
The maximum over $s$ in this formula implies that, contrary to OGH curves, the
optimal solution here cannot be expressed as a single analytical formula on
$\lambda$ and $\mu$. However, the optimum to a relaxation of this problem can.
We call \emph{Hermite curves with Overall Uniformly-Bounded Accelerations}
(HOUBA) the polynomials given by:
\begin{align}
    \lambda^*_\HOUBA & = 6 \cdot \frac{3 (\Delta \cdot \bfv_0) (\bfv_1 \cdot \bfv_1) - 2
    (\Delta \cdot \bfv_1) (\bfv_0 \cdot \bfv_1)}{9 \| \bfv_0 \|^2 \|\bfv_1\|^2
    - 4 (\bfv_0 \cdot \bfv_1)^2} \label{houba1} \\
    \mu^*_\HOUBA & = 6 \cdot \frac{-2 (\Delta \cdot \bfv_0) (\bfv_0 \cdot \bfv_1) + 3
    (\Delta \bfv_1) \| \bfv_0\|^2}{9 \|\bfv_0\|^2 \|\bfv_1\|^2 - 4(\bfv_0 \cdot
    \bfv_1)^2} \label{houba2}
\end{align}
(Recall that $\Delta \defeq \bfp_\goal - \bfp_\cur$.) We show in
Appendix~\ref{oubah} that this formula provides a suboptimal upper bound to the
uniform bound~\eqref{eq:UOGH}. As desired, it yields trajectories that are less
prone to sharp boundary accelerations. In what follows, we interpolate all our
COM and swing-foot paths by HOUBA curves.

\subsection{Swing limb synchronization}

COM and swing-foot path retimings are not independent:~contact constraints on
COM accelerations need to be computed using the single-support CWC while the
swing foot is in the air, and the double-support CWC once contact is made. One
way to take this coupling into account is to \emph{synchronize} both path
retimings. Denote by $s$ and $s_\swing$ the indexes of the COM and swing-foot
paths respectively. Synchronization amounts to setting $s_\swing = s /
s_\trans$ and reformulating all foot constraints (limited workspace velocity
and acceleration) on $(\sdd_\swing, \sd_\swing^2)$ as constraints on $(\sdd,
\sd^2)$. This approach has the merit of conciseness and computational
efficiency, but we chose not to do so due to an undesired side effect: at the
end of the swing-foot trajectory, we want the foot velocity to go to zero so as
to avoid impacts. Under synchronization, this would imply that the COM velocity
goes to zero as well, and thus that the contact switch is quasi-static.

To avoid this issue, we adopted the two-stage retiming strategy depicted in
Figure~\ref{fig:synchro}:
\begin{itemize}
    \item First, retime the swing-foot trajectory. Let $T_\swing$ denote the
        duration of the resulting trajectory.
    \item Second, retime the COM trajectory under the additional constraint
        that $t(s_\trans) > T_\swing$, \ie that the retimed COM trajectory will
        spend \emph{at least} $T_\swing$ of its time in the single-support
        section $s \in [0, s_\trans]$.
\end{itemize}
Given the initial path velocity $\sd_0 = \|\pd_\cur\| / \|\ps(0)\|$, the
constraint $t(s_\trans) > T_\swing$ can be formulated as a path acceleration
constraint $\sdd \leq \sddmax$ suitable for TOPP:

\begin{prop}
    \label{prop:sddmax}
    A sufficient condition for $t(s_\trans) > T_\swing$ is that $\sdd \leq
    \sddmax$, with
    \begin{equation}
        \sddmax \ \defeq \ \frac{1}{2 s_\trans}\left[
            \left(\frac{s_\trans}{T_\swing}\right)^2 - \sd_0^2 \right]
        \label{eq:sddmax}
    \end{equation}
\end{prop}
The proof of this property is given in Appendix~\ref{app:sddmax}.
In the $(\ddot s,\dot s^2)$-plane where TOPP-polygons are computed,
Equation~\eqref{eq:sddmax} amounts to adding two vertical boundaries at $\sdd =
\pm \sddmax$ and can be done readily in existing software.

\subsection{Inverse kinematics controller}

The last step of our pipeline converts COM and foot reference accelerations
$\pdd_G, \pdd_\swing$ into joint commands that are finally sent to the robot's
motor controllers. We used our own inverse kinematics solver for this, which is
implemented in the open-source \emph{pymanoid}
library.\footnote{\url{https://github.com/stephane-caron/pymanoid}} The solver
is based on a single-layer quadratic program with weighted whole-body tasks
(see~\cite{caron2017zmp} for details). We used the following tasks, by
decreasing priority: support foot contact, swing foot tracking, COM tracking,
constant angular-momentum and joint-velocity minimization, with respective
weights $10^4, 100, 10, 1$. We chose joint-velocity minimization as the
regularizing task. Each iteration of the solver updates a vector of joint-angle
velocities $\qd_\textrm{ref}$ which is finally sent to the robot.

\section{Experiments}
\label{sec:experiments}

\begin{figure*}
    \includegraphics[width=0.95\textwidth]{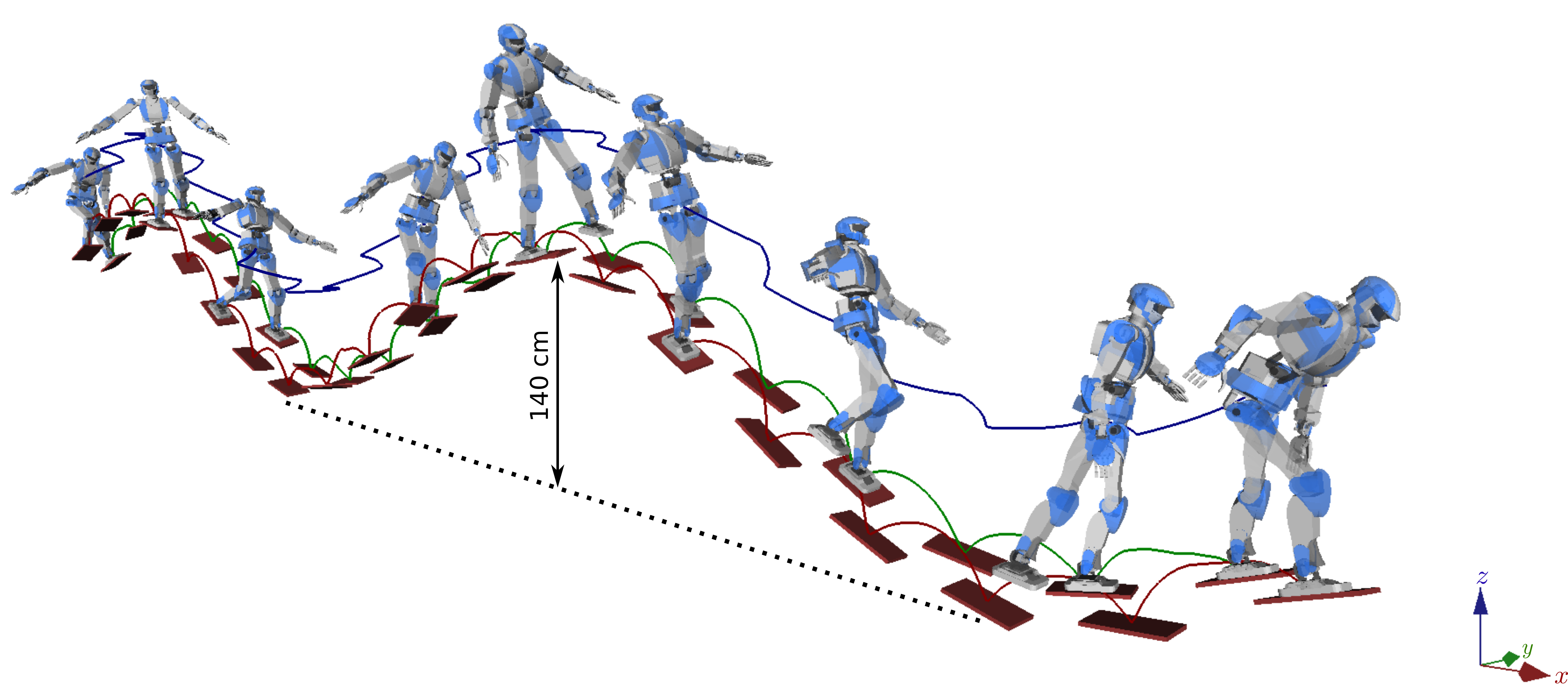}
    \caption{
        \textbf{Locomotion over uneven terrain by HRP-4 running the TOPP-MPC
        controller.} The only external input to the system is a foothold
        sequence. From this, the controller automatically deduces the
        \emph{timings} of its single- and double-support phases (which depend
        on terrain topology), along with COM and swing-foot motions that
        guarantee contact stability. TOPP previews are run in a closed feedback
        loop with an update period of 40~ms. In the simulations depicted above,
        the robot climbs up and down two meter-high hills with slopes ranging
        from $0^{\circ}$ to $30^{\circ}$. Feasibility of the motion has been
        checked by computing at each time instant feasible supporting contact
        forces, which can be seen in the accompanying video.
    }
    \label{fig:walking}
\end{figure*}

We evaluated TOPP-MPC in simultations with a model of the HRP-4 humanoid robot.
All the source code necessary to reproduce our work is publicly
released~\footnote{\url{https://github.com/stephane-caron/topp-mpc}}.

\subsection{Preview targets tuning}

The main input required by our controller is a sequence of foothold locations,
which we assume to be provided by a parallel perception-and-planning module.
From these footholds, target COM and swing-foot positions and velocities are
computed as follows. Let $(\bft_i, \bfb_i, \bfn_i)$ denote the frame of the
$\th{i}$ contact in the sequence. For the DS phase ending on this contact and
the SS phase of the $\th{(i - 1)}$ contact, preview targets are set to:
\begin{itemize}
    \item $\bfp_G^\goal$ is the center of the SEP of the next SS footstep
    \item $\pd_G^\goal = v_{\textrm{ref}}\,\bft_i$, where $v_{\textrm{ref}} = 0.4$~m.s$^{-1}$
    \item $\bfp_\swing^\goal$ is the position of the $\th{i}$ contact
    \item $\pd_\swing^\goal = \alpha \bft_i - (1 - \alpha) \bfn_i$, where
        $\alpha$ tunes the forward inclination of the foot landing velocity.
\end{itemize}
Similarly, we used $\bfv_0 = \beta \bft_{i-1} + (1 - \beta) \bfn_{i-1}$ for the
foot takeoff direction. In the accompanying video, we set $\alpha=0.5$ and
$\beta=0.3$. Recall that only the signs and directions of these velocity
vectors matter as we use Hermite curves for interpolation.

\subsection{TOPP tuning}

Once a path is interpolated, we retime it using a path discretization step of
$\d{s} = 0.1$. Minimum-time trajectories always saturate inequality
constraints, which means in particular that output contact wrenches $\bfw_G$
will lie on the boundary of the CWC, a behavior that would make our controller
sensitive to disturbances. We palliate this by computing the CWC matrix with
sole dimensions and friction coefficients downscaled by $0.75$.

\subsection{Simulations}

Figure~\ref{fig:walking} shows one application of our controller on a scenario
where the robot has to climb up and down a series of meter-high hills, with
slopes ranging from 0$^\circ$ to $30^\circ$. In this example, swing-foot
accelerations were bounded by $\|\pdd_\swing\| \leq 5$~m.s$^{-2}$. Phase
timings derived by the controller were, on average over the complete motion,
$0.88 \pm 0.06$~s for DS phases (these rather long durations may result from
our choice of COM targets centered in the SEP), and $0.65 \pm 0.22$~s for SS
phases. Note that these phase durations \emph{result from} the particular
terrain topology selected for the experiment.

\begin{table}[h]
    \caption{TOPP performance in the closed MPC feedback loop.}
    \centering
    \small
    \begin{tabular}{ccc}
        Phase & Convex Hull & Bretl \& Lall \\
        \hline   
        DS & 18.6 $\pm$ 5.5\,ms & 26.2 $\pm$ 7.0\,ms \\
        SS & 30.0 $\pm$ 5.6\,ms & 38.9 $\pm$ 7.5\,ms \\
        \hline   
    \end{tabular}
    \label{tab:topp-mpc}
\end{table}

Table~\ref{tab:topp-mpc} reports times taken to build and solve TOPP instances
in this simulation. We observe that the convex-hull reduction presented in
Section~\ref{reduction} yields results $20$ to $30\%$ faster than when using
Bretl and Lall's method. Overall, TOPP-MPC fits in an update loop running at
30~Hz.

\section{Conclusion}
\label{sec:conclusion}

We introduced a model predictive controller whose underlying optimization is a
Time-Optimal Path Parameterization (TOPP-MPC). Despite its nonlinearity, we
showed how the TOPP optimization runs fast enough for the control loop in
legged locomotion. The key feature of TOPP-MPC is that it determines by itself
proper timings between contact switches, based on terrain topology and its
model of system dynamics. We evaluated the performance of the overall
controller in simulations where the humanoid model HRP-4 climbs up and down a
series of hills (Figure~\ref{fig:walking} and accompanying video).

There are many avenues for further improvements. By construction, time-optimal
retiming switches between maximum and minimum accelerations, causing
discontinuities in the ensuing contact forces. To mitigate this effect, an
active line of research seeks to extend the underlying TOPP routine so as to
enforce continuity constraints on accelerations~\cite{singh2015class} or jerk
bounds~\cite{hung2016structure}. Another direction would be to rely on
admissible velocity propagation~\cite{pham2017admissible} to discover feasible
motions throughout multiple contact changes.

\bibliographystyle{IEEEtran}
\bibliography{refs}

\appendices

\section{Calculation of HOUBA parameters}
\label{oubah}

In this problem, we seek to minimize an upper-bound $M$ and the accelerations
$\|H''(s)\|^2$ of the curve. By definition of Hermite curves,
\begin{eqnarray}
    2 H''(0) & = &  3\Delta - 2 \lambda \bfv_0 - \mu \bfv_1 \label{min1} \\
    2 H''(1) & = & -3\Delta + \lambda \bfv_0 + 2 \mu \bfv_1 \label{min2}
\end{eqnarray}
By convexity of the cost function $\|H''(s)\|^2$, extrema are realized at the
boundaries $s \in \{0, 1\}$ of the interval: $\forall s \in [0, 1],
\|H''(s)\|^2 \leq \max(\|H''(0)\|^2, \|H''(1)\|^2)$. Our problem is therefore
the joint minimization of \eqref{min1}-\eqref{min2}.
By Minkowski inequality, 
\begin{eqnarray}
    2 \|H''(0)\|^2 & \leq & \|3 \Delta - \lambda \bfv_0 - \mu \bfv_1\|^2 +
    \|\mu \bfv_1\|^2 \\
    2 \|H''(1)\|^2 & \leq & \|3 \Delta - \lambda \bfv_0 - \mu \bfv_1\|^2 +
    \|\lambda \bfv_0\|^2
\end{eqnarray}
Using the symmetry in $\lambda$ and $\mu$, we reformulate this as the
minimization of $E(\lambda, \mu) := \|3 \Delta - \lambda \bfv_0 - \mu
\bfv_1\|^2 + \frac12 \lambda \|\bfv_0\|^2 + \frac12 \mu \|\bfv_1\|^2$.
Differentiating with respect to $\lambda$ and $\mu$ yields:
\begin{eqnarray}
    {\partial E}/{\partial \lambda} & = & -6 (\Delta \cdot \bfv_0) + 9
    \lambda \|\bfv_0\|^2 + 6 \mu (\bfv_0 \cdot \bfv_1) \label{dE1} \\
    {\partial E}/{\partial \mu} & = & -6 (\Delta \cdot \bfv_1) + 6 \lambda
    (\bfv_0 \cdot \bfv_1) + 9 \mu \| \bfv_1 \|^2 \label{dE2} 
\end{eqnarray}
Finally, solving for critical points the linear system given by $\eqref{dE1}=0$,
$\eqref{dE2}=0$ yields the formulas~\eqref{houba1} and~\eqref{houba2}.

\section{Proof of Property~\ref{prop:sddmax}}
\label{app:sddmax}

Let us define $a(s) \defeq \sdd$, $b(s) \defeq \sd^2$, and denote by $b'(s)
\defeq \frac{\d{b}}{\d{s}}$ and $\dot{b}(s) \defeq \frac{\d{b}}{\d{t}}$. The
definitions of $a$ and $b$ imply that $b'(s) = 2 a(s)$, so that
\begin{equation}
    \textstyle
    b(s) \ = \ \sd_0^2 + \int_0^{s} b'(s) \d{s}
    \ = \ \sd_0^2 + 2 \int_0^{s} a(s) \d{s}.
\end{equation}
An upper bound $a(s) \leq \sddmax$ then implies that
\begin{equation}
    b(s_\trans) \ \leq \ \sd_0^2 + 2 s_\trans \sddmax.
\end{equation}
Next, the output switching time $t(s_\trans)$ can be written as:
\begin{equation}
    t(s_\trans) \ = \ \int_0^{s_\trans} \frac{\d{s}}{\sqrt{b(s)}}
    \ \geq \ \frac{s_\trans}{\sqrt{\sd_0^2 + 2 s_\trans \sddmax}}
\end{equation}
A necessary condition for $t(s_\trans) > T_\swing$ is thus that $s_\trans^2
\geq T_\swing^2 (\sd_0^2 + 2 s_\trans \sddmax)$. Equation~\eqref{eq:sddmax}
is finally a rewriting of the latter inequality.

\end{document}